\documentclass[letterpaper]{article}
\usepackage{aaai}

\usepackage{amsmath,amssymb,amsfonts}
\usepackage[ruled,vlined]{algorithm2e}
\usepackage{graphicx}
\usepackage{subfig}
\usepackage{svg}
\usepackage[utf8]{inputenc}
\usepackage{textcomp}
\usepackage{xcolor}
\usepackage{url}

\usepackage{times}
\usepackage{helvet}
\usepackage{courier}

\begin{document}

\title{Co-generation of game levels and game-playing agents\\
}

\author{Aaron Dharna \\ Dept. of Computer and Information Science \\ Fordham University \\ aadharna@gmail.com \And
Julian Togelius \\ Tandon School of Engineering \\ New York University \\ julian@togelius.com \And
L. B. Soros \\ Tandon School of Engineering \\ New York University \\ lsoros@nyu.edu}

\maketitle

\begin{abstract}
Open-endedness, a longstanding cornerstone of artificial life research, is the ability of systems to generate potentially unbounded ontologies of increasing novelty and complexity. Engineering generative systems displaying at least some degree of this ability is a goal with clear applications to procedural content generation in games. The Paired Open-Ended Trailblazer (POET) algorithm, heretofore explored only in a biped walking domain, is a coevolutionary system that simultaneously generates environments and agents that can solve them. This paper introduces a POET-Inspired Neuroevolutionary System for KreativitY (PINSKY) in games, which co-generates levels for multiple video games and agents that play them. This system leverages the General Video Game Artificial Intelligence (GVGAI) framework to enable co-generation of levels and agents for the 2D Atari-style games Zelda and Solar Fox.
Results demonstrate the ability of PINSKY to generate curricula of
game levels, opening up a promising new avenue for research at the intersection of procedural content generation and artificial life. At the same time, results in these challenging game domains highlight the limitations of the current algorithm and opportunities for improvement.

\end{abstract}


\section{Introduction}
Humans acquire skills incrementally, e.g. learning to crawl before learning to walk. In this way, primitive skills serve as building blocks for more complex and difficult behaviors. Curricula, which are sets of related tasks or increasingly complicated versions of the same task, can scaffold incremental acquisition of skills for hard problems that are too complex to solve from scratch.
However, AI research does not often make use of curricula, instead operating on largely unrelated task domains. 
Yet, curriculum generation is an important problem because although deep learning algorithms and other recent innovations have achieved landmark performance on historically unsurmounted benchmark domains such as the board game Go \cite{SilverGo2016} (which has long served as a grand challenge for artificial intelligence) and the video game Montezuma's Revenge \cite{ecoffet2019goexplore}, the challenge of developing intelligent processes that perform well \emph{in general} remains unmet. 


The recent Paired Open-Ended Trailblazer (POET) algorithm \cite{Wang:gecco19} took an initial step towards open-ended curriculum generation by evolving parameters for a 2D biped locomotion domain (i.e. hill slope and obstacle placement) while simultaneously evolving agent controllers. \emph{Open-ended} connotes the scale of creativity seen in biological evolution \cite{stanley:oreilly2017}. POET's coevolutionary system was able to generate unique adaptive curricula for learning to walk on uneven terrain. However, it is unknown what other kinds of curricula can be generated coevolutionarily. Games are rich domains for exploring this question  because they require critical skills not necessary for bipedal walking, such as long-term planning to avoid enemies. However, many standard video-game-based reinforcement learning (RL) benchmarks
are unsuitable for curriculum generation because the games cannot be modified.

This paper describes a novel system called PINSKY that co-generates gameplay agents and levels for games in the General Video Game AI (GVGAI) competition framework. 
First, necessary background on procedural content generation is reviewed and the POET algorithm is described in full detail. The PINSKY system is then introduced and key differences from original POET (necessary for generating and playing games) are explicitly noted. Results show that PINSKY can co-generate levels and agents for the 2D Zelda- and Solar-Fox-inspired GVGAI games, automatically evolving a diverse array of intelligent behaviors from a single simple agent and game level. Though there are limitations to level complexity and agent behaviors, our analysis suggests reasons for these limitations and directions for future research.

\section{Background}


Early work on gameplay AI centered around tree search methods such as A* and minimax~\cite{yannakakis2018artificial}. 
The successive development and wider adoption of artificial neural networks (ANNs) allowed further innovation for game-playing AI. Methods for optimizing ANNs generally fall into five categories: supervised learning, unsupervised learning, RL, evolutionary approaches, and hybrid learning approaches \cite{justesen:tog20}. RL approaches to gameplay in particular generally involve an agent interacting with an environment and repeatedly gaining some amount of reward for its actions.
Learning, then, is an optimization process that maximizes long-term reward. Modern RL systems have achieved success in part by incorporating \emph{self-play} (which can be viewed as a form of coevolution [\citeauthor{Arulkumaran_2019} \citeyear{Arulkumaran_2019}]) into learning schemes, at least for two-player competitive games. In this paradigm, policies being learned are played against each other, with the resulting gameplay data then affecting the trajectory of the learning algorithm. Recent examples of high-performing self-play systems include AlphaGo \cite{SilverGo2016}, AlphaStar \cite{Vinyals2019AStar}, and OpenAI Five \cite{openai2019dota}, though, notably, these systems all require human gameplay data for initial bootstrapping. 
\citeauthor{justesen2018illuminating} (\citeyear{justesen2018illuminating}) 
demonstrate that \emph{automatically} generating levels at an agent-appropriate difficulty level dramatically improves performance on 2D games. However, the level generators for the four games in their study (adapted versions of Zelda, Solar Fox, Frogger, and Boulder Dash) incorporated human-designed elements specific to each game,
thereby both demonstrating the utility of generating training levels on-the-fly and 
highlighting that there is still a need for truly human-free level generation systems.

\subsection{Search-based procedural content generation}


Procedural Content Generation (PCG) refers to a variety of methods for algorithmically creating novel artifacts, from static assets such as art and music to game levels and mechanics. Much research is devoted to creating levels that provide adequate challenge and could have plausibly been created by a human level designer. 
Importantly, the work described in this paper is not focused on creating levels plausibly created by human designers. Instead, it creates game levels that a) satisfy specific playability constraints, b) increase in complexity over time, and c) coevolve alongside algorithmically-controlled game-playing agents.

\emph{Search-based} PCG in particular has been theorized to potentially lead to truly endless games~\cite{TogeliusSearch2011}. The search-based approach requires three primary components: 1) a search algorithm, 2) a content representation, and 3) an evaluation function \cite{shaker2016procedural}. The search algorithm component of such systems frequently (but not always) takes the form of
an evolutionary algorithm, wherein a population of content artifacts is created and gradually varied in order to maximize an evaluation function. For the purpose of creating games, the evaluation function can incorporate information from automated gameplay\cite{TogeliusSearch2011}. For example, playtraces can be examined for lead changes~\cite{browne2010evolutionary}, the capacity of an agent to learn to play the game can be measured~\cite{togelius2008experiment}, or the performance of several agents on a game can be compared~\cite{nielsen2015general}. 
In any case, methods designed \emph{only} to optimize objective-oriented fitness metrics can result in incomplete or stagnated search \cite{lehman:ec2011}. In contrast, the work reported in this paper sees one realization of endlessly creating diverse levels for game-playing agents to learn from where the levels grow more complex over time.





\subsection{The POET Algorithm}
The Paired Open-Ended Trailblazer (POET) algorithm \cite{Wang:gecco19} is a coevolutionary system for concurrently generating and solving new environments. The approach first explored the OpenAI Gym's Hardcore Bipedal Walker domain, wherein environments consist of obstacle-laden hills. Given rangefinder sensors and joint angle information, agents must learn gaits that allow them to walk far over difficult terrain. POET coevolves agents and terrains through three main processes: 1) periodically generating new environments by mutating existing parents, 2) incrementally optimizing agents paired with environments, and 3) occasionally attempting to transfer optimized agents into new environments. An overview is given in Algorithm \ref{alg:gvgaiPOET}:

\begin{algorithm}[ht]
\SetAlgoLined
 Pair initial environment with unoptimized agent\\
 \While{not done}{
  \If{ counter \% mutationTimer == 0}{
    Generate offspring environment-agent pairs\\ 
    Remove too-easy and too-difficult offspring \\
    \If{population size exceeded}{Remove oldest environment-agent pairs}
  }
  
  Perform a fixed number of optimization steps\\
  Reevaluate all optimized individuals 
  
  \If{counter \% transferTimer == 0}{
    Evaluate all agents on all environments \\
    Replace incumbent agents with more successful agents, if any exist
  }
  counter += 1
 }
 \caption{POET Algorithm}
 \label{alg:gvgaiPOET}
\end{algorithm}

Importantly, generated environments must satisfy a minimal criterion (for viability) that the level is neither too easy nor too hard. The reward function for biped walkers is continuous, allowing ``neither too easy nor to hard'' to be defined by minimal and maximal acceptable reward values. This binary approach to fitness, explored recently in the context of artificial life and evolutionary robotics \cite{LehmanMC}, presents a potentially more open-ended alternative to
traditional gradient-based evolution. After an environment satisfies the difficulty criteria, it inherits a copy of its parent's neural 
controller. Another important and unusual feature of POET is that it periodically evaluates all possible pairs of agents and environments in the population, thereby
revealing behaviors that can be easily adapted to multiple environments. Experiments showed that such \emph{transfers} are necessary for solving difficult walking problems. 
Through incremental optimization and regular transfer of agents, POET generates viable curricula for biped walking. 

It should be noted upfront that POET's dynamics are still largely unknown, as few experiments have actually ever been performed. In fact, 
\citeauthor{Wang:gecco19} (\citeyear{Wang:gecco19}) reported  results from \emph{only three runs} because of the high computational cost. 



\section{Methodology}

This section primarily describes the novel PINSKY system\footnote{code: \url{tinyurl.com/ydgf64wa}}, which adapts the POET algorithm to generating game levels and gameplay agents instead of biped walkers and terrains. 
PINSKY is composed of three interacting subsystems: 1) the GVGAI game framework, 2) an evolutionary level generator, and 3) an incremental game-playing agent optimizer.

\begin{itemize}
\item \textbf{GVGAI Framework:} The General Video Game Artificial Intelligence (GVGAI) framework \cite{perezliebana2018general} provides an interface for defining and playing games written in Video Game Description Language (VGDL), which is a text language for 2D games and levels ranging from dungeon crawlers and RPGs to platformers. Two GVGAI games, Zelda and Solar Fox, are explored in this paper. The GVGAI framework affords multiple tracks of interaction with the games including automated level generation and gameplaying. PINSKY uses both capabilities in tandem to build a population of agent-environment pairs that coevolve over time such that the game levels become more complex while the agents become more proficient (i.e. solving these increasingly complex tasks).

\item \textbf{Evolutionary Level Generator:} Environment evolution in PINSKY begins with a seed level from which all future levels descend. New offspring levels are generated by mutating tiles on a copy of the parent map.
There are three types of possible map mutations, each with separate probabilities: 1) removing a non-player sprite, 
2) adding a new sprite, or 3) moving an existing sprite. 
After each mutation is performed, there is a 50\% chance of another mutation occurring. Ultimately, the new map is deemed viable if it can pass a minimal playability criterion check (described later in this section) whereupon the new agent-map pair inherits its parent's neural network and joins the population of actively optimizing environments. However, the new agent-map pair does not replace its parent in the population.

\item \textbf{Incremental Gameplay Agent Optimizer:} Gameplay agents are controlled by fixed-topology convolutional neural networks, depicted in Figure \ref{fig:TileNet}, reducing the problem of finding good agents to a search through connection parameter space. When an agent-offspring pair is initially created via mutation, the offspring agent is an exact copy of the parent agent. Note that, as in the original POET algorithm, optimization occurs incrementally with a fixed number of optimization steps being executed during each main algorithm loop to adapt offspring networks to their new environments. Preliminary experiments investigated a variety of optimizers, including REINFORCE, PPO \cite{schulman2017proximal}, a simple ES, CMA-ES \cite{Hansen06thecma}, OpenAI's ES \cite{salimans2017:ESRL}, PEPG \cite{sehnkePEPG}, and Differential Evolution (DE) \cite{StornDE}. DE, a population-based optimizer, was selected because of its good convergence properties, ease of implementation, parallelizability, and scalability to high-dimensional problems. 
\end{itemize}



\subsection{Differences from POET}

\subsubsection{Games add complexity and diversity}

The range of game types that even a single game domain can encompass is immense. For example, in dZelda the
task is to pick up a key and take it to the exit while staying alive. However, given a flexible representation (such as VGDL), the game can trivially be changed into a ``connect the dots'' game wherein the agent must pick up a key and then find a path connecting all doors. The win conditions of these two possible dZelda varieties are vastly different, highlighting the future potential for generating arbitrary games. 
Furthermore, 
games inherently
enable more complex behaviors than traditional evolutionary robotics domains because winning frequently involves interacting nontrivially with other agents.



\begin{figure}[bp!]
  \centering
  \includegraphics[width=0.45\textwidth]{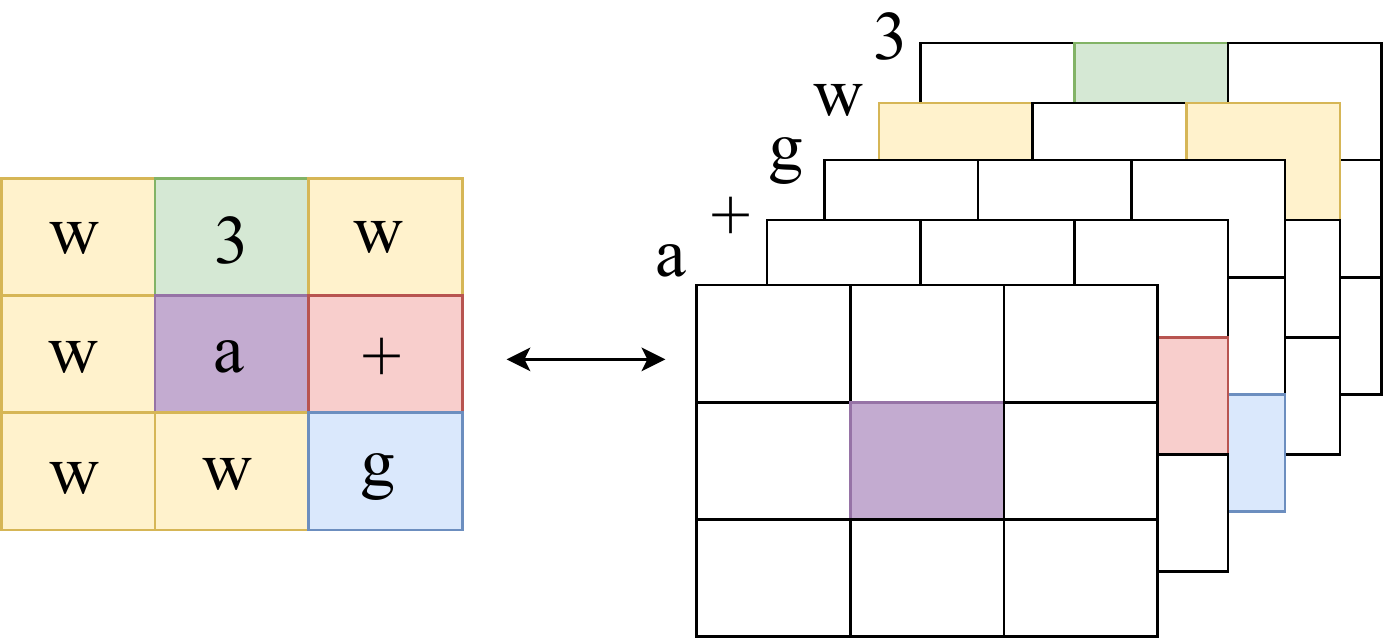}
  \caption{\textbf{One-hot encoded map input to the convolutional policy network.} Tiles in each GVGAI map (left) correspond to $x$,$y$ positions in the environment. In this example from dZelda, possible tiles include (w)all, (g)oal, (a)vatar, key (+), and monster (3). These 2D maps are then extended into a tensor (right) where each slice denotes the presence (indicated by color) or absence of each tile type.}
  \label{fig:mapTensor}
\end{figure}

\subsubsection{ANN Input}
\begin{figure}[bp!]
  \centering
  \includegraphics[width=0.45\textwidth]{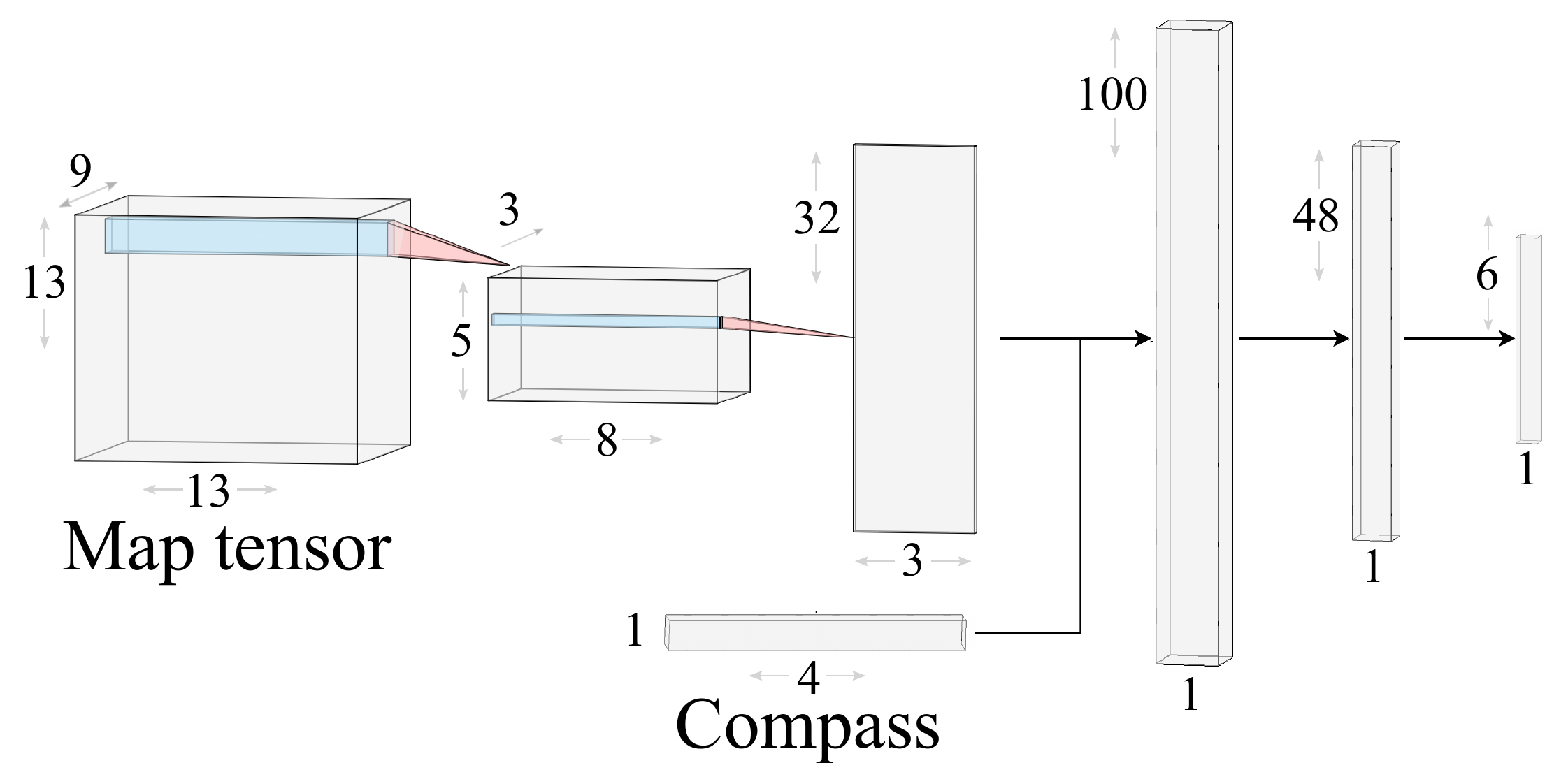}
  \caption{\textbf{Dual-input convolutional policy network for dZelda.} As input, the network takes both the one-hot encoded GVGAI maps (Figure \ref{fig:mapTensor}) and agent orientation information, then produces an action. The Solarfox network structure is minimally different because the map tensor is a different shape (length 11, width 10, depth 14) and the agent can select from fewer actions.}
  \label{fig:TileNet}
\end{figure}

The POET agent had access to rangefinder readings and information about its own joint angles. In this agent-centric paradigm, each action results only from local state information.
In contrast, PINSKY agents are given a tile map of the environment as input to their neural networks (Figures \ref{fig:mapTensor} and \ref{fig:TileNet}) in addition to the agent's orientation. Giving the agent access to global game state and local agent state information allows for more complicated behaviors to emerge. Furthermore, moving away from purely agent-centric network inputs enables the potential generalization of PINSKY to arbitrary games, as most 2D Atari-style games can arguably be represented as some sort of tile map.
A benefit of this new tile input is that it reduces the policy network size. Having fewer parameters makes available evolutionary optimization methods that previously were incapable of training policy networks due to not scaling well. 

\subsubsection{Reward Function}

The RL problem of credit assignment, or determining which actions cause the observed outcome, are hard even when the reward function is dense. A sparse function makes this task even more difficult.
Games such as Solarfox and dZelda are substantially more difficult than the reward-dense biped walker domain.
In Solarfox, the goal task of picking up coins is the only reward source. The dZelda agent is rewarded for picking up a  key, taking it to the door (the win condition), and killing monsters. Solely killing monsters can also earn more reward than winning the game, providing a distracting reward. Reward in both games can be sparse because only specific behaviors earn reward.  



\subsubsection{Minimum Playability Criteria}

POET prevents evolutionary search from degenerating by requiring that evolved terrains satisfy a minimal criterion (MC) \cite{LehmanMC} defined \textit{a priori}; the walker had to be able to walk at least a minimum amount (ensuring the level is not too hard) and at most a maximum amount (ensuring the level is not too easy). In PINSKY, the minimal criterion concept has been adapted into a playability criterion. Specifically, a level is too easy if a random agent \textit{can} beat the level and too hard if a Monte Carlo Tree Search agent (with the default GVGAI time limit of 40ms of planning time per action) \textit{cannot} beat the level. Methods such as MCTS are limiting because having a fast forward model is often an onerous requirement. Furthermore, even with a fast forward model, planning algorithms like MCTS are still subject to variable performance \cite{Nelson2016MCTS}. Nevertheless,  MCTS is robust enough to function as a playability check that can solve a variety of complex levels for these particular games. 

The MC combined with age-based culling allows evolutionary drift to introduce new challenges that the neural network agents will coevolve with. The random mutation in the Evolutionary Level Generator is biased towards adding new objects (e.g. enemies) into the levels, disrupting existing policies. Culling by age provides ample time for the entire population of agents to attempt to solve the new task through direct optimization of the paired agent and repeated transfer attempts of all other agents to replace the paired agent.



\subsection{Experiments}

As a reminder, the initial POET experiments consisted of three runs in a single biped walking domain. Three PINSKY experiments are similarly performed, however each run explores a different game domain and thereby highlights unique capabilities of the novel system. Experimental parameters are in Table \ref{table:GVGAIPOETarguments}. While such a small number of runs precludes statistically significant analysis, demonstrating the viability of this new approach to co-generating game levels and gameplaying agents \emph{at all} despite significant computational limitations is worthwhile in its own right.

\begin{table}[t]
  \centering
    \begin{tabular}{ |p{2cm}|p{1.0cm}|p{3.9cm}|  }
    \hline
    Argument & Default & Description\\
      \hline \hline
  game          & dZelda & GVGAI game to play\\
  gameLen       & 500    & Max actions per game\\ 
  nGames        & 1500   & DE evals per opt. step\\  
  popSize       & 50     & DE population size\\
  mutationTimer & 25     & Loops before mutation step\\
  maxChildren   & 8      & Max offspring per parent\\
  mutationRate  & 0.8    & Parent level mutation\\
  transferTimer & 10     & Loops until transfer attempt\\
  maxEnvs       & 30     & Agent-env. pair pop. size\\
  numPoetLoops  & 5000   & Max PINSKY loops\\
 \hline
    \end{tabular} \caption{PINSKY parameters}
    \label{table:GVGAIPOETarguments}
\end{table}

The first two experiments demonstrate PINSKY performance on two dZelda variants. The first dZelda experiment type (singleDoor) permits only single-door environments, wherein game complexity is increased by adding and rearranging enemies, walls, and keys. The second dZelda experiment type (multiDoor) additionally permits multiple doors in each level, subtly transforming the game from a relatively simple dungeon crawler 
into a more complex game requiring planning to take one key to \textit{all} doors within the time constraints. All dZelda experiments start with the same seed level (Figure \ref{fig:singledoormaps}, left). 

The third experiment type demonstrates the broad generative potential of PINSKY by additionally investigating the GVGAI game Solarfox. Solarfox differs from dZelda in terms of enemy behaviors; while dZelda enemies move freely and kill on direct contact, Solarfox enemies (exactly two per level) can only move around the level's perimeter, but have projectile attacks. The generated neural networks for playing Solarfox have a slightly different topology than dZelda networks; the tile representation includes a separate sheet for the second enemy character, and the set of actions the agent can take does not include combat, therefore fewer output nodes are required. 
Furthermore, Solarfox operates on a different movement scheme than the tile-based movement of dZelda. Movement in Solarfox is continuous, where the agent moves in millimeters in the game. In that case, the tile representation discretizes the space into tiles. 
Because the Solarfox agent requires many more moves to cross the map than the dZelda agent longer games were needed to ensure the minimum playability criterion could be met (so that MCTS reliably solves human-designed levels).


The potentially open-ended nature of POET-like systems means that each run of the algorithm could, in theory, continue forever. However, practical constraints on computational resources necessarily limit runs. In the original POET experiments, each run lasted 10 days in wall clock time \cite{Wang:gecco19} while harnessing 256 parallel CPU Cores (with no mention of RAM). The 
experiments reported herein ran on a 32-core CPU using 50 GB of RAM per experiment.  

\section{Results}

Table \ref{table:GVGAIPOETstats} contains summary statistics for the initial runs. 
While multiDoor dZelda ran for a full 5000 loops, the other runs were truncated to free up computational resources.
Specifically, the singleDoor dZelda run was terminated once it displayed substantial generative potential so the Solarfox run could begin. For the purpose of investigating PINSKY on complex domains at longer timescales, multiDoor dZelda was allowed to complete its full 5000 main algorithm loops. 

\begin{figure}[t]
    \centering
    \subfloat[singleDoor dZelda from seed to solved environment]{\includegraphics[width=0.45\textwidth]{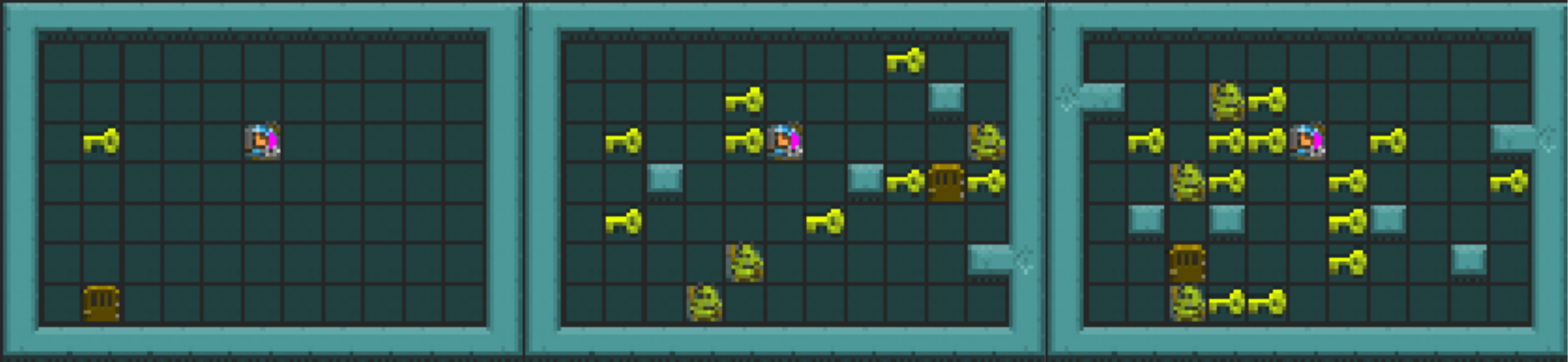}\label{fig:singledoormaps}}
    \hfill
    \subfloat[multiDoor dZelda lineage snapshots]{\includegraphics[width=0.45\textwidth]{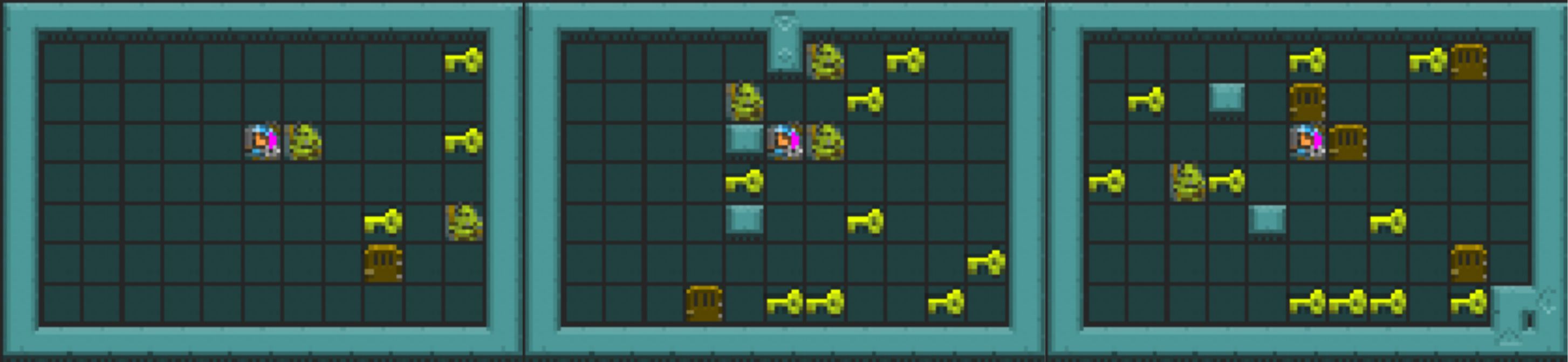}\label{fig:multidoormaps}}
    \hfill
    \subfloat[Solarfox lineage snapshots]{\includegraphics[width=0.45\textwidth]{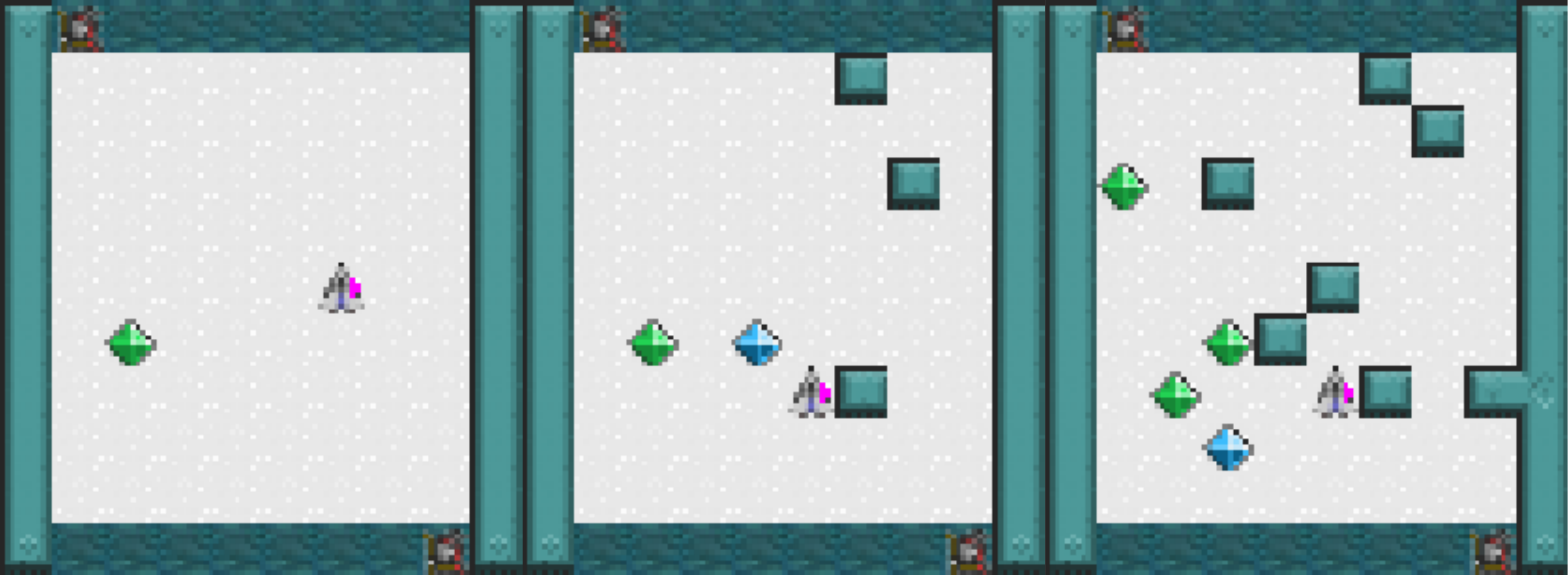}\label{fig:solarfoxmaps}}
    \caption{Lineage Snapshots across 3 PINSKY experiments with native GVGAI reward schemes, showing increasing difficulty over evolutionary time.}\label{fig:snaps}
\end{figure}

Figure \ref{fig:snaps} depicts lineages of generated levels, with task complexity increasing over evolutionary time. Successful dZelda agents tend to follow shortest-distance paths measured in Manhattan distance. Of course, not all behaviors are efficient or even effective. Consider an example policy observed on a level similar to the seed level (Figure \ref{fig:singledoormaps}, left). The degenerate agent takes the key, moves one step, swings its sword to kill the monster, then keeps swinging forever. 

When playing Solarfox (which has a sparse but non-distracting reward signal), PINSKY agents  solve 84.8\% of generated levels that passed the playability criterion. For comparison, 64\% of playable singleDoor dZelda levels and only 12.7\% of multiDoor dZelda levels were solved. Therefore, two additional dZelda singleDoor experiments were run for 5000 loops each using a non-distracting, or \emph{aligned}, reward function that encourages efficient solutions: 

\vspace{-3.5mm}

\[R = \begin{cases} 
      1 - \frac{n_{steps}}{gameLen} & \text{agent reaches goal} \\
      \frac{n_{steps}}{gameLen} - 1 & \text{agent dies} \\
      0 & \text{agent neither reaches goal nor dies} 
   \end{cases}
\]

PINSKY generated 1512 and 1251 viable levels and concurrently solved 90\% and 83\% of viable levels, respectively, which is comparable to performance on Solarfox. Similarly, when multiDoor dZelda uses the non-distracting reward function, 1344 viable levels were generated of which 29\% (compared to 12.7\% previously) were solved.

The minimal playability criterion requires that all generated levels have a MCTS solution before an agent-level pair can be added to the PINSKY population. It is interesting, then, to note that most, but not all, generated multiDoor dZelda levels remain \emph{unsolved}. The rightmost level in Figure \ref{fig:multidoormaps}, generated relatively late in its lineage, is an example \emph{solved} level. The agent takes an efficient path: down to a key, up to the door above its starting position, down and right to the nearby door, up to the right corner door, and then down to the bottom right door. The agent that solves the rightmost Solarfox level in Figure \ref{fig:solarfoxmaps} immediately begins moving left (to avoid crashing into the wall) until it is between the three clustered coins and has cleared the second wall fragment. Once there, it moves down to pick up the bottom-most coin, back up to pick up the upper coin, and then farther left to pick up the third coin in the small cluster. Finally, it continues left until it is partially below the final coin and then moves up to the final coin, thereby ending the game.


\begin{table}[th!]
  \centering
    \begin{tabular}{ |p{2.55cm}|p{1.0cm}|p{1.4cm}|p{1.2cm} | }
    \hline
    Statistic & dZelda & multiDoor & Solarfox\\
      \hline \hline
    Duration  & 8 days & 15 days & 7 days\\
    Loops / & 2411 & 5000 & 3300\\
    Generations &  &  & \\
    Total levels     & 768 & 1600 & 1056\\
    Viable levels    & 684  & 1353 & 448\\
    Solved levels      & 443 & 173 & 380\\
    Transfer attempts & 216900 & 450000 & 297000\\
    Transfers    & 3705 & 8560 & 730\\
    
 \hline
    \end{tabular} \caption{PINSKY results across three different domains.}
    \label{table:GVGAIPOETstats}
\end{table}

The claim that levels become harder over time is validated with a curriculum extraction process from prior work by \citeauthor{wang2020enhanced} (\citeyear{wang2020enhanced}). For each experiment, one lineage leading to a solved level was identified. Solved levels from the lineage’s first 10\%, middle 45-55\%, and last 10\% were then randomly selected. The late-stage levels were optimized with DE from scratch and given the same number of rollouts as PINSKY. Direct DE optimization fails to solve late-stage dZelda levels (verifying that they are nontrivially difficult), but \emph{does} solve ``hard'' Solarfox levels. The selected levels were then concatenated into a curriculum, and ANNs were optimized with DE to sequentially solve the easy, medium, and then hard levels. The agent received the same amount of optimization time as PINSKY does in each environment. Results (Table \ref{table:curriculumLearning}) indicate that the ability to solve difficult levels eventually tapers off even with the behavioral scaffolding of a curriculum. This result reconfirms the findings of \citeauthor{wang2020enhanced} that transferring agents from their original environments into new ones is critical for POET-like systems. 



\begin{table}[tbp]
  \centering
    \begin{tabular}{ |p{3.2cm}|p{1cm}|p{1.4cm}|p{1cm} | }
    \hline
    Experiment & Easy & Medium & Hard\\
    \hline \hline
    Solarfox          & \checkmark & \checkmark & \checkmark\\
    singleDoor        & \checkmark & \checkmark & X\\
    singleDoor aligned 1   & \checkmark & X & X\\
    singleDoor aligned 2   & \checkmark & X & X\\
    multiDoor         & \checkmark & X & X\\
    multiDoor aligned & \checkmark & X & X\\

 \hline
    \end{tabular} \caption{\textbf{Levels solved using an extracted curriculum}: From each sample lineage (Figure \ref{fig:snaps}) a \emph{PINSKY-solved} easy, medium, and hard level were randomly picked to form a curriculum as in Enhanced POET \cite{wang2020enhanced}. Then optimization was performed using the extracted curriculum, validating the necessity of transfer to solve difficult levels. 
    }
    \label{table:curriculumLearning}
\end{table}


\section{Discussion}

The results in the previous section show that even with a small curriculum, agents cannot be optimized to solve harder levels independent of the larger PINSKY algorithm. This result then begs an intriguing question: which components enable solving hard levels? Inspecting data from the original PINSKY runs reveals that successful agents were frequently transferred from levels they were not initially paired with, highlighting the importance of periodic transfer attempts in this coevolutionary system. However, the system still cannot find agents that generalize to solve \emph{all} generated levels, suggesting that more core algorithm innovation is needed.

Evaluating PINSKY on domains with different reward schemes additionally reveals important insights for designing POET-like systems. In particular, the alignment of the reward function with the task to be solved dramatically affects agent performance.
In dZelda, points are earned for completing the primary goal, but also for semi-related subtasks such as killing monsters. Given enough time, the agent maximizes its score by exclusively completing distracting subtasks. 
Inversely, if a game has a non-distracting reward signal (like bipedal walking), then PINSKY functions more like POET.

Over time, PINSKY tends to converge with respect to solvability. To illustrate this phenomenon, consider a dZelda level where the agent starts next to monsters. Agents can (and do) solve such levels, but only via highly specific actions, i.e. instantly turning and attacking. Finding good ANN weights then becomes a search for a needle in a haystack. As levels of appropriate difficulty become rarer in the population, the relative optimization step frequency increases, allowing new levels to be created sooner. However, because 1) level mutations add complexity more than removing it, 2) agents are continually optimized, and 3) older (i.e. simpler) agent-environment pairs are cut from the population before newer ones, there is little incentive for evolving easier levels.

Despite the potential for convergence, the results in the previous section demonstrate that PINSKY is capable of co-generating lineages of increasingly complex game levels and agents that can play them. That being said, the generated levels are visibly different from human designs. For example, PINSKY rarely builds contiguous walls. This particular idiosyncracy could be mitigated in an ad hoc manner by modifying the evolutionary mutation operators. However, it is interesting for the sake of building open-ended generative systems to consider more bottom-up and domain-agnostic incentives for meaningful design. One possible way to rectify this situation might instead focus on increasing generalizability of agent behaviors; although agents are evaluated on multiple domains when domain transfers are attempted, the system doesn't explicitly reward solving multiple levels. 

It is possible that adding more demanding incentives could encourage the evolution of more challenging environments. This discussion raises the question of why we should even bother generating ANNs when tree search algorithms can already solve the types of Atari-style games explored in this paper. However, using tree search agents would limit the system to domains with a fast forward model available, excluding most interesting scenarios. The pursuit of generalizable gameplay agents is also worthwhile in its own right, and PINSKY may prove to be a useful tool in this regard. Ideally, the combination of incremental agent optimization with periodic transfer of agents to new environments will result in agents not having time to overfit to their respective environments, which is a phenomenon commonly observed in deep RL \cite{cobbe2018quantifying}. However, for the current approach to be truly successful, network architectures or training methods that generalize better will likely need to be devised.

The pursuit of open-ended evolutionary and generative processes has long been a goal of artificial life research, and the experiments reported in this paper suggest that much can be learned from cross-pollination between these historically disconnected fields. For instance, experiments in a virtual evolving world show that manipulating the minimal viability criterion can speed or slow evolution \cite{Soros2016}. Similarly adjusting the viability criterion in a POET-like system would be interesting from an evolutionary dynamics perspective because of the complex interactions between the level generator and the optimizer. It should additionally be noted that the GVGAI framework explicitly makes possible the evolution of game \emph{mechanics},
offering another promising avenue for future work with PINSKY.

\section{Conclusion}
This paper adapted the coevolutionary POET algorithm to simultaneously generating game levels and agents that can solve them. Adapting the algorithm to games from its original bipedal walker domain required innovations with respect to key differences from the original algorithm and domain, including enabling more complex environments, giving new kinds of information to gameplay agent controllers, and having an extremely sparse reward function. Results on a limited number of runs demonstrate that the system can, in fact, be adapted to co-generate game levels and game-playing agents while nonetheless illuminating future directions for making the generated levels both more difficult and more solvable. However, it appears that the failure of the trained deep networks to generalize 
cannot be overcome only by transferring agents from one game level to another.


\section*{Acknowledgements}
This work was supported by the National Science Foundation. (Award number 1717324 - “RI: Small: General Intelligence through Algorithm Invention and Selection.”). 

\bibliographystyle{aaai}
\bibliography{bibliography}

\end{document}